\documentclass{bmvc2k}

\usepackage{graphicx}
\usepackage{booktabs}
\usepackage{multirow}
\usepackage{wrapfig}
\usepackage[autostyle]{csquotes}
\usepackage{pifont}

\usepackage[accsupp]{axessibility} 
\usepackage[vlined,linesnumbered,ruled]{algorithm2e}
\usepackage{orcidlink}

\title{CosFairNet:A Parameter-Space based Approach for Bias Free Learning}

\addauthor{Rajeev Ranjan Dwivedi}{rajeev22@iiserb.ac.in}{1}
\addauthor{Priyadarshini Kumari}{priyadarshini.kumari@sony.com}{2}
\addauthor{Vinod K Kurmi}{vinodkk@iiserb.ac.in}{1}

\addinstitution{
 Department of Data Science and Engineering\\
Indian Institute of Science Education and Research Bhopal\\
India
}
\addinstitution{
Sony AI, USA
}

\runninghead{Rajeev R Dwivedi, Priyadarshini K, Vinod K Kurmi}{CosFairNet}

\def\etal{\emph{et al}\bmvaOneDot}

\begin{document}

\maketitle

\begin{abstract}
Deep neural networks trained on biased data often inadvertently learn unintended inference rules, particularly when labels are strongly correlated with biased features. Existing bias mitigation methods typically involve either a) predefining bias types and enforcing them as prior knowledge or b) reweighting training samples to emphasize bias-conflicting samples over bias-aligned samples. However, both strategies address bias indirectly in the feature or sample space, with no control over learned weights, making it difficult to control the bias propagation across different layers. Based on this observation, we introduce a novel approach to address bias directly in the model's parameter space, preventing its propagation across layers. Our method involves training two models: a bias model for biased features and a debias model for unbiased details, guided by the bias model. We enforce dissimilarity in the debias model's later layers and similarity in its initial layers with the bias model, ensuring it learns unbiased low-level features without adopting biased high-level abstractions. By incorporating this explicit constraint during training, our approach shows enhanced classification accuracy and debiasing effectiveness across various synthetic and real-world datasets of different sizes. Moreover, the proposed method demonstrates robustness across different bias types and percentages of biased samples in the training data. The code is available at: \href{https://visdomlab.github.io/CosFairNet/}{https://visdomlab.github.io/CosFairNet/}
\end{abstract}

%-------------------------------------------------------------------------
% %\vspace{-0.6cm}
\section{Introduction}
%\vspace{-0.3cm}
\label{sec:intro}
% %\vspace{-1em}
In recent times, there has been a growing concern about the latent possibilities of growing bias and fairness issues within artificial intelligence (AI), deep learning frameworks and models. Bias can infiltrate AI frameworks during various stages, starting from data acquisition, extending to model creation and algorithm development, and even up to the deployment stage~\cite{torra}. The increasing use of deep learning models in various sensitive and high-impact applications makes it extremely crucial to identify and mitigate bias at all potential stages to ensure the development of fair and trustworthy AI workflows~\cite{mehrabi2021survey}. 

Deep learning models tend to learn easy-to-learn features and attributes much faster than hard-to-learn features such as actual shapes and high-level abstraction of an object \cite{nam2020learning, geirhos2018imagenet}. For instance, while training a cow classifier, the model may fail to classify the cow if it is placed on the surface of a lake or on a beach. This happens because the model has a contextual bias, and to make a confident prediction, the correct context, typically a green grassland, is required. The absence of a non-correlating background or context results in incorrect prediction. The deep networks are prone to noise memorization faster and quicker than their intended purpose~\cite{arpit2017closer}. The models often suppress shapes and learn color~\cite{dosovitskiy2014discriminative}, texture~\cite{geirhos2018imagenet} and attribute bias~\cite{hooker2020characterising}. Henceforth, the model is dependent on bias and so performs better on in-bias or bias-aligned samples and fails to perform as soon as there is a bias shift or as the non-correlated data samples are encountered. A classic example of this is a cow in a green pasture and a camel in a desert. If the backgrounds of these images are exchanged, the model's performance drops drastically~\cite{beery2018recognition}. These unintended bias leads the trained model to make erroneous inferences by relying on shortcuts.
\begin{wrapfigure}{b}{0.38\textwidth}
\centering
    \includegraphics[width=0.38\textwidth]{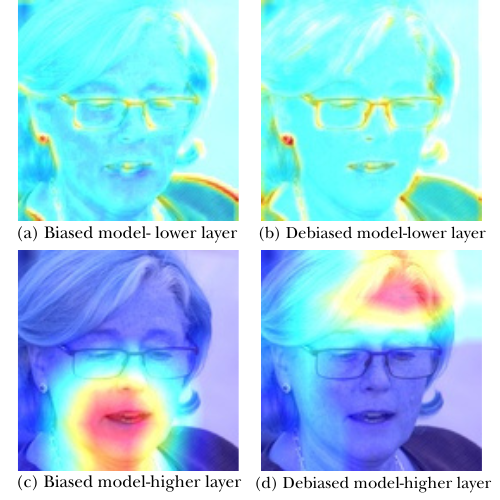}
    % \small
\caption{ \smaller Gradient activation map of biased and debiased models in lower and higher layers for the BFFHQ dataset. (\textbf{a}) and (\textbf{b}) illustrate  gender predictions of the biased and debiased models at the initial layer. It is evident that the debiased model closely corresponds to the biased model. In contrast, (\textbf{c}) and (\textbf{d}) show no correspondence as the learned weights differ at the higher layer. (Best view in color). }
    \label{fig:teaser}
\end{wrapfigure}

 Current methods for bias removal use various approaches. A substantial body of research relies on explicit bias annotations~\cite{kim2019learning, li2019repair, sagawa2019distributionally, teney2020unshuffling, krueger2021out, bai2021decaug}. However, this reliance on explicit annotations can be costly and time-consuming, necessitating a thorough understanding of the potential spurious correlations between bias attributes and target labels. Another significant area of study in bias mitigation centres on algorithms that utilize reweighting techniques~\cite{lee2021learning, nam2020learning}. These algorithms assign higher weights to the bias-conflicting samples than the bias-aligned samples during the training phase. A notable approach, among reweighting techniques, involves training two models: (a) a bias model that learns the dataset's bias and (b) a debias model that leverages the learning of the bias model to identify bias-aligned samples and weigh them down during training phase so that a spurious correlation between bias attributes and the target label is not formed. Subsequently, in the continuation to the reweighting, several studies propose improvement by modifying the model architecture ~\cite{Lee2022,kim2022learning}. 

Although the proposed methods demonstrate their effectiveness in specific applications, they encounter some major challenges. While one type of method suffers the challenge of expressing and quantifying biases precisely for the model to handle them explicitly, another type based on sampling or re-weighting is prone to assigning disproportionately high weights to noisy or outlier samples, especially when they constitute a minority in the dataset. Furthermore, when down-weighting bias-aligned samples, these techniques also discard valuable debias features alongside the bias features, which could benefit the learning process. Moreover, relying solely on the existing re-weighting technique may introduce new spurious correlations between bias and target labels when multiple biases exist.

In this study, we address some of the existing challenges by introducing a novel method that mitigates bias in the model's parameter space. It is observed that the learned parameters of a biased model and a debiased model at initial layers are similar, while the parameters in the final stage layers differ. To illustrate this observation, we generate the gradient activation maps~\cite{selvaraju2017grad} for predictions from biased and debiased models at both initial and later layers. In Fig.~\ref{fig:teaser}, (a) and (b) demonstrate that both models at the initial layer utilize common features for gender prediction. In contrast, in (c) and (d), the parameters of the biased model in the later stage utilize the mouth region, while the debiased model employs other features for gender identification. Our method builds on the insights that low-level features of biased samples are not detrimental to learning, but it is the unintended correlation between biased features and target labels that poses the problem. Thus, instead of discarding low-level features of bias-aligned samples by simply down-weighting them, we propose to harness them by enforcing a (dis)similarity constraint in the parameter space of the debias and bias model. Specifically, we enforce similarity constraints on the initial layers of these models while introducing an orthogonality constraint on the final layers. The purpose of these constraints is twofold: first, to ensure the preservation of low-level features from all input samples, including bias-aligned ones, while simultaneously preventing the formation of unintended correlations with the target label. The orthogonality constraint, applied at the final layer, compels the debias model to focus on learning the signal rather than the bias. By applying this constraint in the parameter space, our method also prevents the propagation of biases through subsequent layers. Our main technical contributions are:
\begin{itemize}
    \item We introduce a novel approach to mitigate bias by model parameter realignment, along with a unique architecture design to prevent the acquisition and propagation of bias during model training. 
    \item We demonstrate the \textit{utility of bias-aligned samples} and propose to leverage them in the model training through a simple yet effective constraint within the model's parameter space. Further, by applying an orthogonality constraint to the later-stage layers, we direct the debias model to acquire distinct learning compared to the bias model, thus preventing spurious correlations between bias attributes and target labels.
    \item The proposed method demonstrates superior performance when compared to both types of approaches: those relying on bias labels and those solely based on sample re-weighting schemes across two real-world datasets and two well-controlled synthetic datasets. 
\end{itemize}

\section{Related Work}
This section classifies prior research in the field of debiasing techniques, with a specific emphasis on reducing bias. Debiasing efforts can be categorised into two primary groups: debiasing through the utilisation of prior knowledge and debiasing approaches based on reweighting. These categories include a range of methods, each with its own distinct methodology and consequences, which are evaluated and compared to our approach.

 \textbf{Debiasing using prior knowledge}
A variety of works focus on mitigating bias through prior knowledge of bias types in datasets, utilizing bias annotations and distinct prediction heads for different biases \cite{wang2019learning, tartaglione2021end, bahng2020learning, kim2022learning, sagawa2019distributionally, teney2020unshuffling, krueger2021out, Kurmi_2021_aai_gba, bai2021decaug, li2023whac}. Some methods, like those proposed by Wang \etal~\cite{wang2019learning} and End \etal~\cite{tartaglione2021end}, employ predefined representations of bias or orthogonal gradients to debias data, though these can be complex and cost-intensive to apply in real-world settings.

 \textbf{Debiasing using sample reweighting}
Reweighting-based debiasing methods involve sample weighting, label usage, or model architectures to counter biases \cite{kim2019learning, li2019repair, nam2020learning, lee2021learning, vandenhirtz2023signal, qraitem2023bias, Lee2022, kim2022learning, jeon2022conservative}. Techniques range from dual model architectures to ensemble methods and feature-level swapping. However, methods like Bias Swap \cite{kim2021biaswap} or those by Wu \etal~\cite{wu2023discover}, which rely on image translation, are limited by their complexity. Other approaches, such as those using adversarial learning or alternative loss functions \cite{bahng2020learning, wang2019learning, amplibias}, show promise but often perform similarly to conventional models.

 In summary, existing debiasing techniques vary widely, from employing prior knowledge to reweighting samples and using complex model architectures. Despite their innovative approaches, these methods often face challenges related to implementation complexity, costs, and the handling of hyperparameters. Our novel approach aims to synthesise the strengths of these methodologies while overcoming their limitations to improve debiasing effectiveness.

\section{Problem Formulation}

We address this as a \( \mathcal{N} \)-class classification problem where a dataset \( \mathcal{D} \) contains inputs \( x \) with attributes \( \{a_1, \ldots, a_k\} \), each attribute \( a_i \) taking values from a predefined set \( A_i \). The main objective is to construct a predictive function \( f \) that operates under a specific set of decision rules \( \mathcal{F}_t \). This function \( f \) aims to accurately predict the target attribute \( y \), where \( y = a_t \) and \( a_t \) is an element of the set \( A_t \). For the purpose of determining whether or not there is bias within the dataset \( \mathcal{D} \), the following conditions are crucial:
\begin{enumerate}
    \item \textbf{Correlation Condition:} There exists a non-target attribute \( a_b \), different from \( y \), that correlates significantly with \( y \). This is quantitatively assessed using the conditional entropy \( H(y | a_b) \), which approaches zero, indicating that \( a_b \) nearly determines \( y \).
    \item \textbf{Decision Rule Condition:} There is an effective alternative decision rule \( s_b \), not included in \( \mathcal{F}_t \), capable of classifying based on \( a_b \). This suggests that \( a_b \) could independently serve as a strong predictor for \( y \).
\end{enumerate}

\noindent When these conditions are met, \(a_b \) is recognised as a bias attribute in \(\mathcal{D} \). Furthermore, instances in \(\mathcal{D} \) are classified according to their alignment with the bias attribute \(a_b \). \textbf{Bias-Aligned:} An instance is considered bias-aligned if it conforms to the alternative decision rule \(s_b \), which align with the bias observed in \(a_b \). \textbf{Bias-Conflicting:} Bias-conflicting instances are those that do not conform to \(s_b \), indicating non-typical behaviour despite a correlation between \(a_b \) and \(y \). The development of \( f \) and \( \mathcal{F}_t \) must not only achieve high prediction accuracy for \( y \) but also address and mitigate the biases associated with \( a_b \) to ensure fairness and equity in the predictive outcomes across all classes in \( \mathcal{D} \).

\begin{wrapfigure}{!}{0.5\textwidth}
%\vspace{-0.9cm}
\centering
    \includegraphics[width=0.5\textwidth]{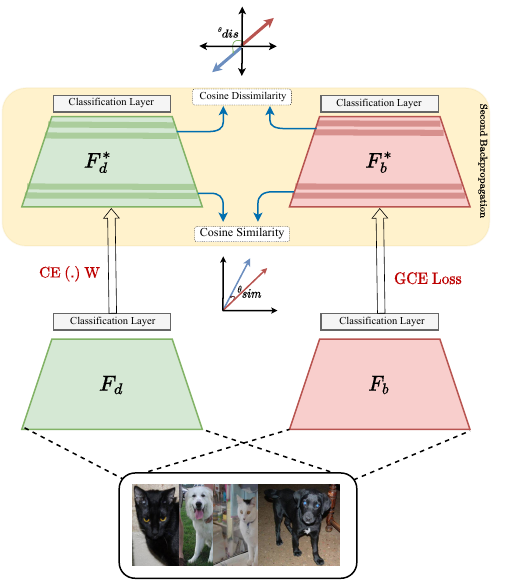}
    \caption{\small The architecture of CosFairNet depicts the debiasing mechanism where \( F_d \) and \( F_b \) are the debias and bias models, respectively. \textit{W} represents the weighting of samples. Cosine similarity is employed to align (initial layers) or de-align (later layers) \( F_d \) and \( F_b \) model layers to ensure differentiated learning of biased and unbiased representations. In figure, $CE(.)$ stands for cross-entropy loss, $GCE(.)$ for generalized cross-entropy loss and $W$ for relative difficulty score. }
    \label{fig:cosfairnet}
\end{wrapfigure}

%\vspace{-0.3cm}
\section{Proposed Method}
%\vspace{-0.3cm}
We propose a novel framework \textbf{\textit{CosFairNet}}, that addresses the problem of bias through a principled realignment of model weights. Our approach is predicated on the use of cosine (dis)similarity measures to modulate the influence of bias, which manifests at varying magnitudes across different network layers. Specifically, we apply cosine similarity to synchronize the lower layers of the bias model \( F_b \) with the corresponding layers of the debias model \( F_d \), thus aligning their basic low-level feature learning. This step ensures that both models extract similar foundational details from the data. In contrast, we employ cosine dissimilarity at the last stage layers to encourage \( F_d \) to diverge from \( F_b \) in its high-level feature representations. During training, \( F_b \) excels in identifying bias-aligned samples. Leveraging this, we make the initial layers of \( F_b \) akin to those of \( F_d \); at the same time \( F_b \) seeks to capture biases, and \( F_d \) aims to learn genuine features. This adversarial arrangement prompts realignment of weights that enhances the performance of both models by prioritizing features that improve \( F_d \)'s debiasing capability. Furthermore, by ensuring the model layers are constrained to the unit hyper-sphere, we measure similarity and dissimilarity using angular distance, defined as follows: 
% $\textit{S} = \mathcal{L}_{cosSim}(F_b, F_d) = \frac{\mathcal{L}_{F_b} \cdot \mathcal{L}_{F_d}}{\|\mathcal{L}_{F_b}\| \cdot \|\mathcal{L}_{F_d}\|}$
\begin{equation}
\textit{S} = \mathcal{L}_{cosSim}(F_b, F_d) = \frac{\mathcal{L}_{F_b} \cdot \mathcal{L}_{F_d}}{\|\mathcal{L}_{F_b}\| \cdot \|\mathcal{L}_{F_d}\|},
\label{eq:cosine_loss}
\end{equation}

\noindent where \( \mathcal{L}_{F_b} \) and \( \mathcal{L}_{F_d} \) are the layer parameters. %, the dot product is denoted by (\( \cdot \)), and the norm by \( \| \cdot \| \).
Given the unit norm constraint, the dot product similarity yields the cosine of the angular separation between the two parameter vectors, with \( S \) ranging from 1 to -1, indicating 0-degree (aligned) to 180-degree (opposed) separations, respectively.
% %\vspace{-1em}
%\vspace{-0.2cm} 
\subsection{Model Architecture} %\vspace{-0.25cm}
% %\vspace{-1em}
The model architecture, as depicted in Fig.~\ref{fig:cosfairnet}, is a dual model framework inspired by the two-model setup from~\cite{nam2020learning}. The architecture consists of two parallel networks: the debiasing model \( F_d \) and the biased model \( F_b \), both featuring the same structural design. The training process is bifurcated into two distinct yet interconnected phases. Initially, \( F_b \) is trained using a generalized cross entropy (GCE) loss to effectively capture bias-aligned samples. Concurrently, \( F_d \) is trained with a weighted cross entropy (CE) loss, wherein the weights \( \mathcal{W} \) are adjusted to give precedence to bias-conflicting samples. A pivotal characteristic of this architecture is the synergistic interaction between the corresponding layers of \( F_d \) and \( F_b \). During the second phase of training, the layers \enquote{$k$} of both models undergo a process of alignment or dealignment, contingent upon their position within the model's layer position.\\
% Following is a more detailed explanation of how biased and debiased models are trained.\\

\noindent \textbf{Training a biased Model}
Our approach to training a biased model involves employing the generalized cross entropy (GCE) loss function \cite{zhang2018generalized} to amplify the model's unintentional decision rule. The GCE loss, defined as \( GCE(p(x;\theta_b),y)= \frac{1 - p_y (x;\theta_b)^q}{q} \), uses \( p(x; \theta_b) \) to represent the softmax output, and \( p_y(x; \theta_b) \) as the probability of the target attribute \( y \). The hyperparameter \( q \), in the range \( (0, 1] \), modulates bias amplification, with \( \text{lim}_{q\rightarrow 0}  \frac{1 - p_y ^q}{q} \) mirroring standard cross entropy (CE) loss. Unlike CE loss, GCE loss's gradient, \( \frac{\partial GCE(p, y)}{\partial \theta_b} = p^q_{y} \frac{\partial CE(p, y)}{\partial \theta_b} \), disproportionately weighs samples where the model's prediction strongly aligns with the target. This emphasis on \enquote{easier} samples leads to an augmented bias in the model's learning process compared to CE-trained networks, hence giving us a strong biased model.\\

\noindent \textbf{Training a debiased model}
While concurrently training a biased model as previously described, we also train a debiased model. This involves employing the CE loss with re-weighting based on a relative difficulty score $\mathcal{W}(x)$. The score is formulated as follows: $\mathcal{W}(x) = \frac{\text{CE}(F_b(x),y)}{\text{CE}(F_b(x),y) + \text{CE}(F_d(x),y)}$. Here, \( F_b(x) \) and \( F_d(x) \) represent the softmax outputs of the biased and debiased models, respectively. This score quantifies the degree to which each sample may introduce bias that conflicts or aligns with our observations. Specifically, for bias-aligned samples, where the biased model \( F_b \) incurs a smaller loss as compared to the debiased model \( F_d \) during early training stages, the difficulty score is low, resulting in a smaller weight for training the debiased model. Conversely, for bias-conflicting samples, where the biased model \( F_b \) experiences a larger loss compared to the debiased model \( F_d \), the difficulty score is high (close to 1), leading to a higher weight for training the debias model.
% \begin{equation}
%     \mathcal{W}(x) = \frac{\text{CE}(F_b(x),y)}{\text{CE}(F_b(x),y) + \text{CE}(F_d(x),y)}
%     \label{eq:difficulty_score}
% \end{equation}
% %\vspace{-0.5cm}
\subsection{Realignment of Debias-Model's Parameters }
In the consequent training phase, the \( F_b \) and \( F_d \) models are updated with \(\operatorname{GCE}(p(x;\theta_b); y_i)\) and \(\mathcal{W} \cdot \operatorname{CE}(p(x;\theta_d); y_i)\) respectively. Realignment of \( F_d \) parameters with \( F_b \) occurs within the same batch and step, using cosine similarity as a loss function (see Eq.~\ref{eq:cosine_loss}). This similarity measure ranges from \(-1\) to \(1\), indicating how vectors are oriented in the multidimensional space. During the second back-propagation, cosine similarity and dissimilarity are applied between the models' layers—initial layers use similarity, while later layers use dissimilarity. 

Given \( F_b \)'s ability in correctly classifying bias-aligned samples, the second updates focus solely on \( F_d \)'s layers, leaving \( F_b \) layers unchanged.  It's worth noting that gradients flow from higher to lower layers during backpropagation. As a result, while updating the later layer of the \(F_d \) model, the optimisers may also change the layers before the given layer due to their momentum. To correctly update the \(F_d \) model, all layers must be frozen during the second update except the layer to be updated or their learning rate turned to zero. After the second update of \( F_d \), the training step concludes, enhancing debiasing accuracy. The specifics of this training method are described in the algorithm and can be found in the Supplementary Material.

%\vspace{-0.3cm}
\section{Experiments and Results}
%\vspace{-0.3cm}
% %\vspace{-1em}
We conduct comprehensive experiments to evaluate our method across multiple datasets, including two synthetic datasets: Colored MNIST~\cite{nam2020learning} and Corrupted CIFAR-$10^1$~\cite{nam2020learning} and two real-world datasets:  BFFHQ~\cite{kim2021biaswap} and Dogs \& Cats~\cite{Lee2022}.  We first evaluate the effectiveness of our model for the classification task under different percentages of biased samples within the datasets. Subsequently, we perform additional experiments to address the following research questions:
\begin{enumerate}
\itemsep0em 
\item [Q1] How does the performance of our method change with an increasing ratio of biased samples in the training data, compared to state-of-the-art baselines?
\item [Q2] What is the impact of each constraint, including similarity and dissimilarity, applied to the model's weights on the overall performance?
\item [Q3] How effective are the learned embedded features in separating target classes and distinguishing between bias and debias samples?
\end{enumerate}

\noindent \textbf{Datasets}
We evaluate our debiasing approach using four datasets. The \textbf{Colored MNIST} dataset, a modified version of MNIST with colored digits to induce spurious correlations, ensures consistency in dataset comparisons. The \textbf{Corrupted CIFAR-$10^1$} dataset introduces bias through environmental distortions like fog and brightness changes. The \textbf{BFFHQ dataset}, derived from Flickr-Faces-HQ, uses gender as a bias to analyze age, while the \textbf{Dogs \& Cats dataset} associates animal species with color. Further information regarding the datasets can be found in the supplementary material. Additionally, details and results on the BAR dataset \cite{nam2020learning} can also be found in the supplementary material.

\textbf{Implementation details: }
We use a multi-layer perceptron (MLP) with three hidden layers for Colored MNIST, as suggested by \cite{nam2020learning} and \cite{Lee2022}. For Corrupted CIFAR-$10^1$, we use ResNet20 and train it from scratch with random initialization.  For all other datasets, we use pre-trained ResNet18 \cite{he2016deep} topped with 3 MLP layers with 0.5 dropout between the layers. The Adam optimizer is used for all datasets, with a learning rate of 0.001 for all the datasets. Additionally, we keep the batch size of 256 for Colored MNIST and Corrupted CIFAR-$10^1$, while 64 for the BFFHQ~\cite{kim2021biaswap} and Dogs \& Cats~\cite{Lee2022} dataset. 

\subsection{Q1 - Comparative performance analysis with state-of-the-art under varying bias ratios}
Firstly, we analyze the effectiveness of our model on the well-controlled datasets - Colored MNIST adapted from~\cite{nam2020learning} and Corrupted CIFAR-$10^1$ from~\cite{hendrycks2019benchmarking}, where the bias takes the form of color and corruption, respectively. Table~\ref{table:table1} shows the results on Colored MNIST with different bias-conflicting ratio datasets. \textit{Due to the diverse construction of the synthetic Colored MNIST dataset, we limited our analysis to methods that utilised the same dataset as ours}. As stated by~\cite{flachot2018processing}, the color and small-level features are learned in the early layers, and the information is passed on to the later layers (classification layers) of the model. Therefore, the outcome is expected to be biased in the presence of easy-to-learn bias features, such as color, while intrinsic features like the shape of the digits are suppressed. 

We compare the performance of the proposed model with state-of-the-art methods. Our model consistently outperforms all the methods across all variations of bias ratios.  We also benchmarked our model against the recent work AmpliBias~\cite{amplibias}. As shown in Table~\ref{table:table1}, the proposed model consistently outperforms AmpliBias across all bias ratios by varying margins. Notably, our method demonstrates robust performance, with an average improvement of $3\%$ over LFF and DisEnt, and over $9\%$ compared to AmpliBias at 95\% bias ratio. Similarly, for 99.5\% bias ratio, the proposed model's improvement is 3\% over LFF and DisEnt, and over $1\%$ compared to AmpliBias. 
We suspect that these methods struggle with ineffective sample-reweighting when minority instances are rare, as also stated in~\cite{wu2023discover}. However, we achieve a more balanced performance not only when trained with a very low bias-conflicting ratio but also when the bias-conflicting ratio is significantly high, a scenario often encountered in real-world applications.

\begin{table}[h]
%\vspace{-1em}
    \begin{minipage}[t]{0.43\textwidth}
        \centering
        % \small
        \footnotesize
        \scalebox{0.9}{
        \begin{tabular}{|c|c|c|c|c c|}
            \hline
            & \multicolumn{5}{c|}{\textbf{Colored MNIST} \cite{nam2020learning}} \\ % Center the heading cell
            \hline \hline
            \textbf{Bias Ratio(\%)}&\textbf{99.50} &\textbf{99.00} & \textbf{98.00} & \textbf{95.00}  & \\
            \hline
            Vanilla &  34.75 & 49.87  & 65.72 &  81.72  & \\
            \hline
            HEX \cite{wang2019learning}  &  42.25 & 47.02  & 72.82 & 85.50  & \\
            \hline
            LNL \cite{kim2019learning} &  36.29 & 49.48  & 63.30 &  81.30 & \\
            \hline 
            EnD \cite{tartaglione2021end} &  35.33 &  48.97  & 67.01 &  82.09 & \\
            \hline 
            LFF \cite{nam2020learning} & 63.49 & 72.94 & 80.67 & 85.81 &  \\
            \hline
            DisEnt \cite{vandenhirtz2023signal} & 63.98 & 76.33 &  82.38 &  85.54 &  \\
            \hline
            AmpliBias \cite{amplibias} & 66.01  & 67.79 &  71.32 &  78.88 & \\
            \hline
            BiasEnsemble \cite{Lee2022} & 66.71  & 75.80 &  82.98 &  86.51 & \\
            \hline
            A$^2$ \cite{an20222} &  \textbf{67.47} & 70.68 & 76.93  &   86.09 & \\
            \hline
            \textbf{Ours} & 67.25 & \textbf{78.03} & \textbf{84.22} & \textbf{88.64}  & \\
            \hline 
        \end{tabular} 
        }\vspace{5pt}
        \caption{ \small Comparative accuracy performance (in \%) on bias conflicting samples of various debiasing algorithms on the Colored MNIST dataset, with respect to different bias ratios.}
        \label{table:table1}
    \end{minipage}\hfill
    \begin{minipage}[t]{0.48\textwidth}
        \centering
        \footnotesize
        \scalebox{0.9}{
        \begin{tabular}{|c|c|c|c|c|}
            \hline
            & \multicolumn{4}{c|}{\textbf{Corrupted CIFAR-$10^1$}} \\ % Center the heading cell
            \hline \hline
           \textbf{Bias Ratio(\%)}&\textbf{99.50} &\textbf{99.00} & \textbf{98.00} & \textbf{95.00} \\
            \hline
            Vanilla &  17.93 & 22.72 & 30.21 & 45.24 \\
            \hline
            HEX \cite{wang2019learning}  & 15.39  & 16.62  & 17.81  & 21.74   \\
            \hline
            ReBias \cite{bahng2020learning} & 22.68 & 27.92 & 32.09  & 43.74  \\
            \hline
            EnD \cite{tartaglione2021end} & 20.74 & 24.19 & 38.88 & 40.54  \\
            \hline
            A$^2$ \cite{an20222} & 23.37 & 27.54 & 30.60  &  37.60 \\  
            \hline
            DisEnt \cite{lee2021learning}& 31.97 & 31.22 & 36.98 & 46.40 \\
            \hline
            LFF \cite{nam2020learning}& 29.87 & 33.84 & 40.21  & 51.83 \\
            \hline
            $\epsilon$-SupInfoNCE \cite{barbano2023unbiased} & 33.71  & 38.28 & 41.87 & 51.62  \\
            \hline
            LogitCorrection  \cite{liu2023avoiding} & 34.56 & 37.34 & 47.81  &  54.55 \\  
            \hline
            AmpliBias \cite{amplibias} & 34.63  & \textbf{45.95} &  48.74 & 52.22 \\
            \hline
            \textbf{Ours} & \textbf{36.34} & 43.94 & \textbf{ 50.83} & \textbf{60.06} \\  
            \hline  
        \end{tabular}
        }\vspace{5pt}
        \caption{\small Comparative accuracy performance ( in \%) on bias conflicting samples of various debiasing algorithms on the  Corrupted CIFAR-$10^1$ dataset.}

        \label{table:table2}
    \end{minipage}
    \vspace{1em}
\end{table}

\noindent We tested our method on the CIFAR-10 dataset, which is more complex than the Colored MNIST dataset. As with CMNIST, on the Corrupted CIFAR dataset, the model tends to learn simpler features like textures~\cite{geirhos2018imagenet,arpit2017closer} over intrinsic features. Table~\ref{table:table2} shows our results on the Corrupted CIFAR-$10^1$ dataset. Despite the dataset's difficulty, our method outperforms most current models, including recent advancements that focus on bias mitigation through logit correction loss minimization~\cite{liu2023avoiding}. Introducing orthogonality constraints in the parameter space significantly enhances performance, even when biases such as low-level texture features are present. This improvement may stem from biases in the initial layers not transferring to the final layers, as the debiased model $F_d$ starts similar to $F_b$ but becomes significantly different, leading to an unbiased model.

Our model excels across various bias-conflicting ratios, surpassing most other models~\cite{liu2023avoiding, park2023efficient, barbano2023unbiased}, except at a $99\%$ bias ratio, where Amplibias has a slight $2.01\%$ lead. We believe our model's success is due to its ability to utilize low-level features effectively while preventing false correlations between targets and biases by implementing orthogonality constraints.

In Table ~\ref{table:table4} and Table ~\ref{table:table5}, we present an evaluation of our proposed method against state-of-the-art algorithms on two distinct real-world datasets BFFHQ and Cats $\&$ Dogs respectively. BFFHQ exhibits facial attributes as bias, while Cats $\&$ Dogs involve image bias (color in the image). Similar to synthetic datasets, our method consistently outperforms all baselines, including recent ones, by a significant margin. As presented in Table \ref{table:table4}, our model achieves a notable improvement in accuracy, gaining $12.58\%$ at the $99.00\%$ bias ratio and $7.9\%$ improvement at the $95.00\%$ bias ratio. Moreover, the improvement is substantial (around $35\%$) compared to a vanilla network that does not explicitly address biases. On the BFFHQ dataset \cite{kim2021biaswap}, where age is the intrinsic feature and gender is the bias, our model demonstrates a significant performance boost compared to state-of-the-art methods and the latest studies \cite{amplibias, an20222, barbano2023unbiased}.

\subsection{Q2 - Ablation study on orthogonality constraints} 
Both datasets, BFFHQ and Cats $\&$ Dogs, have distinct features. The significant performance gain across both datasets emphasizes the effectiveness and robustness of our method across diverse features, bias ratios, and scale of the dataset. Although counter-intuitive, our method gives better gains over the vanilla network, on average, in more challenging test scenarios with a severe bias ratio as compared to lesser severe bias scenarios. For instance, our method achieves $45\%$ gain over the vanilla network at $99\%$ bias ratio as compared to $26\%$ gain at $95\%$ bias ratio on Cats $\&$ Dogs dataset, and similarly, a $27\%$ gain at $99.5\%$ bias ratio as compared to $7\%$ gain at $95\%$ bias ratio on BFFHQ dataset.

\begin{table}[h]
% \vspace{-1em}
    \begin{minipage}[t]{0.45\textwidth}
        \centering
        \footnotesize
        \scalebox{0.95}{
        \begin{tabular}{|c|p{1.3cm}|p{1.2cm}|}
            \hline
             \multicolumn{1}{|c|}{} & \multicolumn{2}{c|}{\textbf{ \small Cats \& Dogs Dataset}} \\
            \hline \hline
            \textbf{Bias Ratio(\%)} & \textbf{99.00} & \textbf{95.00} \\
            \hline
            Vanilla & 48.06 & 69.88 \\
            \hline
            HEX \cite{wang2019learning} & 46.76 & 72.60 \\
            \hline
            LNL \cite{kim2019learning} & 50.90 & 73.96 \\
            \hline
            EnD \cite{tartaglione2021end} & 48.56 & 68.24 \\
            \hline
            ReBias \cite{bahng2020learning} & 48.70 & 65.74 \\
            \hline
            LFF \cite{nam2020learning} & 71.72 & 84.32 \\
            \hline
            DisEnt \cite{lee2021learning} & 65.74 & 81.58 \\
            \hline
            BiasEnsemble \cite{Lee2022} & 81.52 & 88.60 \\
            \hline
            \textbf{Ours} & \textbf{93.00} & \textbf{96.50} \\
            \hline 
        \end{tabular}
        }\vspace{5pt}
        \caption{ \small Comparative accuracy performance (in \%) on bias conflicting samples of various debiasing algorithms on the real-world Cat $\&$ Dog dataset.}
        \label{table:table4}
    \end{minipage}\hfill
    \begin{minipage}[t]{0.53\textwidth}
        \centering
        \scriptsize
        \scalebox{1.05}{
        \begin{tabular}{|c|c|c|c|c|}
            \hline
            \multicolumn{1}{|c|}{} & \multicolumn{4}{c|}{\textbf{BFFHQ Dataset}} \\
            \hline \hline
            \textbf{Bias Ratio(\%)} & \textbf{99.50} & \textbf{99.00} & \textbf{98.00} & \textbf{95.00} \\
            \hline
            Vanilla & 55.64 & 60.96 & 69.00 &  82.88 \\
            \hline
            HEX \cite{wang2019learning} & 56.96 &62.32 & 70.72 &  83.40 \\
            \hline
            LNL \cite{kim2019learning} & 56.88 & 62.64 &  69.80 & 83.08 \\
            \hline
            EnD \cite{tartaglione2021end} & 55.96 & 60.88 &  69.72 & 82.88 \\
            \hline
            ReBias \cite{bahng2020learning} & 55.76 & 60.68 & 69.60 &  82.64 \\
            \hline
            LFF \cite{nam2020learning} & 65.19 &  69.24 &  73.08 & 79.80 \\
            \hline
            DisEnt \cite{lee2021learning} & 62.08 & 66.00 & 69.92 &  80.68 \\
            \hline
            BiasEnsemble \cite{Lee2022} & 67.56 & 75.08 & 80.32 &  85.48 \\
            \hline
            A$^2$ \cite{an20222} & 77.83 & 78.98 & 81.13  &  86.22 \\  
            \hline
            AmpliBias \cite{amplibias} & 78.82  & 81.80 &  82.20&  87.34 \\
            \hline
            \textbf{Ours} & \textbf{83.20} & \textbf{82.20} & \textbf{88.40} & \textbf{90.20} \\
            \hline
        \end{tabular}
        }\vspace{5pt}
        \caption{ \small Comparative accuracy performance (in \%) on bias conflicting samples of various debiasing algorithms on the BFFHQ dataset. }
        \label{table:table5}
    \end{minipage}
\end{table}

% \vspace{1em}
\noindent Our method posits that bias in models arises mainly due to spurious correlations between target labels and bias features, which themselves do not inherently cause misclassification but rather contribute to dataset diversity and thus enhance model robustness and generalization. To substantiate this perspective, we performed an ablation study analyzing the effects of various constraints in the weight parameter space, using the CMNIST and Corrupted-CIFAR datasets. Our setup involved two main approaches: initially, we aligned the debias network \(F_d\) with the weights of the biased model \(F_b\) in the early layers; subsequently, we imposed dissimilarity constraints in the later layers of \(F_d\) to further refine the debiasing process.

As indicated in Table~\ref{table:ablation}, applying similarity constraints to preserve bias attributes leads to significant performance gains in both datasets, especially noticeable when the bias sample percentage is at $99\%$. This supports the notion that early neural layers are key in developing robust feature representations to mitigate biases in later stages. Furthermore, imposing dissimilarity constraints on the later-stage layers of models $F_b$ and $F_d$, specifically updating model $F_d$, results in enhanced performance. This aligns with the hypothesis that addressing biases in advanced stages of model training facilitates divergent learning trajectories from the biased model. The introduction of orthogonality constraints assists the debias model in basing decisions on intrinsic features rather than biases, contrasting with the biased model.

\begin{table}[h]
\centering
\footnotesize
\begin{tabular}{|c|c|c|c|c|p{1.2cm}|p{1cm}|}
\hline
\multirow{2}{*}{Method} & \multirow{2}{*}{Dissim} & \multirow{2}{*}{Sim} & \multicolumn{2}{c|}{\textbf{CMNIST}} & \multicolumn{2}{c|}{\textbf{Corrupted CIFAR-$10^1$}} \\ \cline{4-7} 
                        &                      &                      & 99\%           & 95\%          &99\%         & 95\%          \\ \hline
Vanilla                 & \ding{56}           & \ding{56}           & 42.26         & 72.77        & 28.22        & 46.89       \\ \hline
% LFF                     & \texttimes           & \texttimes           & 72.94         & 85.81        & 33.84        & 51.83       \\ \hline
Ours       & \ding{56}          & \ding{52}           & 76.64         & 86.43        & 42.70        & 58.09       \\ \hline
Ours       &     \ding{52}        &     \ding{56}      & 77.89         & 87.81        & 43.40        & 60.05      \\ \hline
Ours             & \ding{52}           & \ding{52}           & \textbf{78.03 }        & \textbf{88.64 }       & \textbf{43.94 }       & \textbf{60.06 }      \\ \hline
\end{tabular}
\caption{\small Ablation study on CMNIST and Corrupted CIFAR-$10^1$ dataset. CosFairNet with similarity(Sim) is applied to the initial layer, while dissimilarity (Dissim) is applied to the later-stage layer. \ding{52} and \ding{56}  indicate \enquote{presence} and \enquote{absence} of a particular constraint, respectively.}
\label{table:ablation}
\end{table}

\subsection{Q3 - Effectiveness of the learned features}  
\begin{wrapfigure}{}{0.4\textwidth}
    \centering
    \includegraphics[scale=0.13]{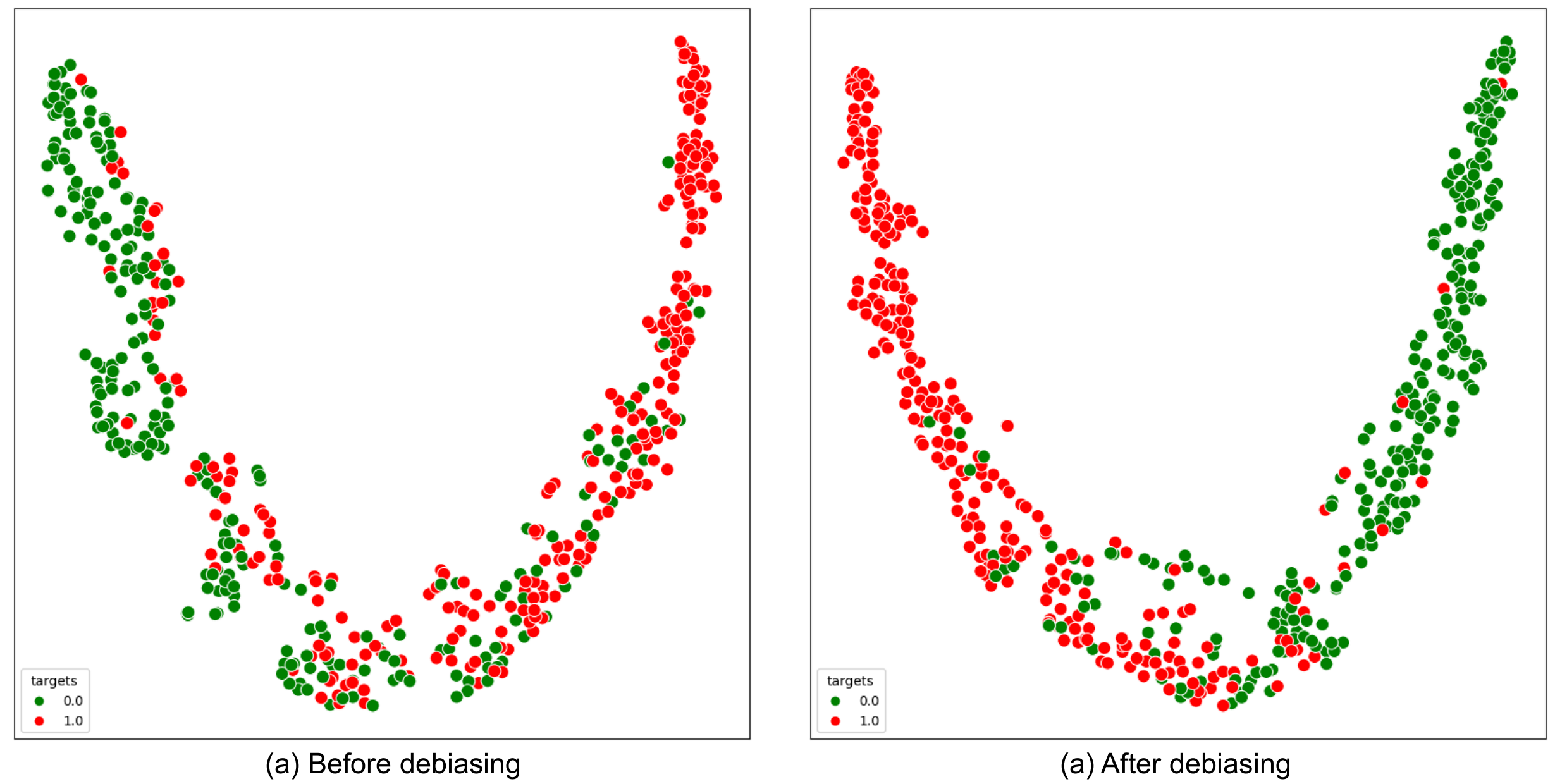}
    \caption{\small t-SNE feature visualization comparing the vanilla model (without debiasing) and the proposed debias model on right side. {\color{red}Red} and {\color{green}Green} points represent samples from two different classes of the BFFHQ dataset. After debiasing with the proposed model, noticeable enhancements in class discrimination are observed, resulting in better separation of classes and hence, better classification accuracy. (Best viewed in color).}
    \label{fig:tnse}
\end{wrapfigure}

To verify our model's ability to learn class-distinguishable features, we used t-SNE embedding on the BFFHQ dataset, which has two target classes. Features from the penultimate layer were visualized in 2D using t-SNE. Fig.~\ref{fig:tnse} displays the differentiation between the features of the vanilla model, which appear mixed, and those of the proposed model, which are distinctly class-separable. This visualization confirms the superior performance of our model, attributable to its effective feature learning.

\section{Discussion and Future Work}
CosFairNet introduces a novel network realignment method to reduce bias, especially when unbiased samples are scarce. This approach builds on the understanding that while low-level features learned in early network layers are not harmful, their spurious correlations with labels can be problematic. By employing an objective function within the parameter space, CosFairNet effectively uses the bias model to enrich these low-level features without creating unwanted correlations between features and labels. Despite its simplicity and efficacy, future work is needed.

 Future directions include identifying which layers are most effective at learning various feature levels, a task complicated by differences in model architecture. For example, complex models may require several initial layers to learn low-level features effectively, whereas simpler datasets might only need the first layer. Another challenge arises when using pre-trained models where the weights are frozen, making it difficult to apply constraints in the parameter space. One potential solution is to introduce additional trainable layers before applying CosFairNet. Furthermore, while the method focuses on bias mitigation, it does not specifically address scenarios with multiple biases for a single class, presenting an opportunity for further exploration and enhancement of the model in handling a broader range of biases.

\section{Conclusions}

In this study, we introduce a novel bias mitigation method that addresses challenges encountered in prior approaches. Leveraging the insight that low-level features of biased samples are valuable for learning, our method focuses on mitigating unintended correlations between biased features and target labels. We utilize the bias model to enhance the learning of more effective and diverse low-level features while employing orthogonality constraints at later-stage layers. The proposed constraints in the parameter space ensure the preservation of low-level features while simultaneously preventing spurious correlations with target labels. Our approach is simple yet effective in mitigating bias and preventing its propagation through subsequent layers, offering a potential solution to some of the limitations of existing methods. In summary, this paper has shown that bias can be effectively curtailed by judiciously adjusting the model's parameters. To tackle bias in machine learning models, our study highlights a novel direction and emphasizes the significance of incorporating diverse feature learning into the training process.

\bibliography{egbib}

\begin{thebibliography}{37}
\providecommand{\natexlab}[1]{#1}
\providecommand{\url}[1]{\texttt{#1}}
\expandafter\ifx\csname urlstyle\endcsname\relax
  \providecommand{\doi}[1]{doi: #1}\else
  \providecommand{\doi}{doi: \begingroup \urlstyle{rm}\Url}\fi

\bibitem[An et~al.(2022)An, Kim, Ko, Lee, and Woo]{an20222}
Jaeju An, Taejune Kim, Donggeun Ko, Sangyup Lee, and Simon~S Woo.
\newblock A\^{} 2: Adaptive augmentation for effectively mitigating dataset bias.
\newblock In \emph{Proceedings of the Asian Conference on Computer Vision}, pages 4077--4092, 2022.

\bibitem[Arpit et~al.(2017)Arpit, Jastrzkebski, Ballas, Krueger, Bengio, Kanwal, Maharaj, Fischer, Courville, Bengio, et~al.]{arpit2017closer}
Devansh Arpit, Stanislaw Jastrzkebski, Nicolas Ballas, David Krueger, Emmanuel Bengio, Maxinder~S Kanwal, Tegan Maharaj, Asja Fischer, Aaron Courville, Yoshua Bengio, et~al.
\newblock A closer look at memorization in deep networks.
\newblock In \emph{International conference on machine learning}, pages 233--242. PMLR, 2017.

\bibitem[Bahng et~al.(2020)Bahng, Chun, Yun, Choo, and Oh]{bahng2020learning}
Hyojin Bahng, Sanghyuk Chun, Sangdoo Yun, Jaegul Choo, and Seong~Joon Oh.
\newblock Learning de-biased representations with biased representations.
\newblock In \emph{International Conference on Machine Learning}, pages 528--539. PMLR, 2020.

\bibitem[Bai et~al.(2021)Bai, Sun, Hong, Zhou, Ye, Ye, Chan, and Li]{bai2021decaug}
Haoyue Bai, Rui Sun, Lanqing Hong, Fengwei Zhou, Nanyang Ye, Han-Jia Ye, S-H~Gary Chan, and Zhenguo Li.
\newblock Decaug: Out-of-distribution generalization via decomposed feature representation and semantic augmentation.
\newblock In \emph{AAAI}, 2021.

\bibitem[Barbano et~al.(2023)Barbano, Dufumier, Tartaglione, Grangetto, and Gori]{barbano2023unbiased}
Carlo~Alberto Barbano, Benoit Dufumier, Enzo Tartaglione, Marco Grangetto, and Pietro Gori.
\newblock Unbiased supervised contrastive learning.
\newblock In \emph{The Eleventh International Conference on Learning Representations}, 2023.
\newblock URL \url{https://openreview.net/forum?id=Ph5cJSfD2XN}.

\bibitem[Beery et~al.(2018)Beery, Van~Horn, and Perona]{beery2018recognition}
Sara Beery, Grant Van~Horn, and Pietro Perona.
\newblock Recognition in terra incognita.
\newblock In \emph{Proceedings of the European conference on computer vision (ECCV)}, pages 456--473, 2018.

\bibitem[Dosovitskiy et~al.(2014)Dosovitskiy, Springenberg, Riedmiller, and Brox]{dosovitskiy2014discriminative}
Alexey Dosovitskiy, Jost~Tobias Springenberg, Martin Riedmiller, and Thomas Brox.
\newblock Discriminative unsupervised feature learning with convolutional neural networks.
\newblock \emph{Advances in neural information processing systems}, 27, 2014.

\bibitem[Flachot and Gegenfurtner(2018)]{flachot2018processing}
Alban Flachot and Karl~R Gegenfurtner.
\newblock Processing of chromatic information in a deep convolutional neural network.
\newblock \emph{JOSA A}, 35\penalty0 (4):\penalty0 B334--B346, 2018.

\bibitem[Geirhos et~al.(2018)Geirhos, Rubisch, Michaelis, Bethge, Wichmann, and Brendel]{geirhos2018imagenet}
Robert Geirhos, Patricia Rubisch, Claudio Michaelis, Matthias Bethge, Felix~A Wichmann, and Wieland Brendel.
\newblock Imagenet-trained cnns are biased towards texture; increasing shape bias improves accuracy and robustness.
\newblock \emph{arXiv preprint arXiv:1811.12231}, 2018.

\bibitem[He et~al.(2016)He, Zhang, Ren, and Sun]{he2016deep}
Kaiming He, Xiangyu Zhang, Shaoqing Ren, and Jian Sun.
\newblock Deep residual learning for image recognition.
\newblock In \emph{Proceedings of the IEEE conference on computer vision and pattern recognition}, pages 770--778, 2016.

\bibitem[Hendrycks and Dietterich(2019)]{hendrycks2019benchmarking}
Dan Hendrycks and Thomas Dietterich.
\newblock Benchmarking neural network robustness to common corruptions and perturbations.
\newblock \emph{arXiv preprint arXiv:1903.12261}, 2019.

\bibitem[Hooker et~al.(2020)Hooker, Moorosi, Clark, Bengio, and Denton]{hooker2020characterising}
Sara Hooker, Nyalleng Moorosi, Gregory Clark, Samy Bengio, and Emily Denton.
\newblock Characterising bias in compressed models.
\newblock \emph{arXiv preprint arXiv:2010.03058}, 2020.

\bibitem[Jeon et~al.(2022)Jeon, Kim, Lee, Kang, and Lee]{jeon2022conservative}
Myeongho Jeon, Daekyung Kim, Woochul Lee, Myungjoo Kang, and Joonseok Lee.
\newblock A conservative approach for unbiased learning on unknown biases.
\newblock In \emph{Proceedings of the IEEE/CVF Conference on Computer Vision and Pattern Recognition}, pages 16752--16760, 2022.

\bibitem[K~Kurmi et~al.(2022)K~Kurmi, Sharma, Sharma, and Namboodiri]{Kurmi_2021_aai_gba}
Vinod K~Kurmi, Rishabh Sharma, Yash~Vardhan Sharma, and Vinay~P. Namboodiri.
\newblock Gradient based activations for accurate bias-free learning.
\newblock In \emph{AAAI,}, Feb 2022.

\bibitem[Kim et~al.(2019)Kim, Kim, Kim, Kim, and Kim]{kim2019learning}
Byungju Kim, Hyunwoo Kim, Kyungsu Kim, Sungjin Kim, and Junmo Kim.
\newblock Learning not to learn: Training deep neural networks with biased data.
\newblock In \emph{Proceedings of the IEEE/CVF Conference on Computer Vision and Pattern Recognition}, pages 9012--9020, 2019.

\bibitem[Kim et~al.(2021)Kim, Lee, and Choo]{kim2021biaswap}
Eungyeup Kim, Jihyeon Lee, and Jaegul Choo.
\newblock Biaswap: Removing dataset bias with bias-tailored swapping augmentation.
\newblock In \emph{Proceedings of the IEEE/CVF International Conference on Computer Vision}, pages 14992--15001, 2021.

\bibitem[Kim et~al.(2022)Kim, Hwang, Ahn, Park, and Kwak]{kim2022learning}
Nayeong Kim, Sehyun Hwang, Sungsoo Ahn, Jaesik Park, and Suha Kwak.
\newblock Learning debiased classifier with biased committee.
\newblock \emph{arXiv preprint arXiv:2206.10843}, 2022.

\bibitem[Ko et~al.(2023)Ko, Lee, Park, Noh, Park, and Kim]{amplibias}
Donggeun Ko, Dongjun Lee, Namjun Park, Kyoungrae Noh, Hyeonjin Park, and Jaekwang Kim.
\newblock Amplibias: Mitigating dataset bias through bias amplification in few-shot learning for generative models.
\newblock In \emph{Proceedings of the 32nd ACM International Conference on Information and Knowledge Management}, CIKM '23, page 4028–4032, New York, NY, USA, 2023. Association for Computing Machinery.
\newblock ISBN 9798400701245.
\newblock \doi{10.1145/3583780.3615184}.
\newblock URL \url{https://doi.org/10.1145/3583780.3615184}.

\bibitem[Krueger et~al.(2021)Krueger, Caballero, Jacobsen, Zhang, Binas, Zhang, Le~Priol, and Courville]{krueger2021out}
David Krueger, Ethan Caballero, Joern-Henrik Jacobsen, Amy Zhang, Jonathan Binas, Dinghuai Zhang, Remi Le~Priol, and Aaron Courville.
\newblock Out-of-distribution generalization via risk extrapolation (rex).
\newblock In \emph{International Conference on Machine Learning}, pages 5815--5826. PMLR, 2021.

\bibitem[Lee et~al.(2021)Lee, Kim, Lee, Lee, and Choo]{lee2021learning}
Jungsoo Lee, Eungyeup Kim, Juyoung Lee, Jihyeon Lee, and Jaegul Choo.
\newblock Learning debiased representation via disentangled feature augmentation.
\newblock \emph{Advances in Neural Information Processing Systems}, 34:\penalty0 25123--25133, 2021.

\bibitem[Lee et~al.(2022)Lee, Park, Kim, Lee, Choi, and Choo]{Lee2022}
Jungsoo Lee, Jeonghoon Park, Daeyoung Kim, Juyoung Lee, Edward Choi, and Jaegul Choo.
\newblock Revisiting the importance of amplifying bias for debiasing.
\newblock \emph{AAAI-23}, 5 2022.
\newblock URL \url{http://arxiv.org/abs/2205.14594}.

\bibitem[Li and Vasconcelos(2019)]{li2019repair}
Yi~Li and Nuno Vasconcelos.
\newblock Repair: Removing representation bias by dataset resampling.
\newblock In \emph{Proceedings of the IEEE/CVF conference on computer vision and pattern recognition}, pages 9572--9581, 2019.

\bibitem[Li et~al.(2023)Li, Evtimov, Gordo, Hazirbas, Hassner, Ferrer, Xu, and Ibrahim]{li2023whac}
Zhiheng Li, Ivan Evtimov, Albert Gordo, Caner Hazirbas, Tal Hassner, Cristian~Canton Ferrer, Chenliang Xu, and Mark Ibrahim.
\newblock A whac-a-mole dilemma: Shortcuts come in multiples where mitigating one amplifies others.
\newblock In \emph{Proceedings of the IEEE/CVF Conference on Computer Vision and Pattern Recognition}, pages 20071--20082, 2023.

\bibitem[Liu et~al.(2023)Liu, Zhang, Sekhar, Wu, Singhal, and Fernandez-Granda]{liu2023avoiding}
Sheng Liu, Xu~Zhang, Nitesh Sekhar, Yue Wu, Prateek Singhal, and Carlos Fernandez-Granda.
\newblock Avoiding spurious correlations via logit correction.
\newblock In \emph{The Eleventh International Conference on Learning Representations}, 2023.
\newblock URL \url{https://openreview.net/forum?id=5BaqCFVh5qL}.

\bibitem[Mehrabi et~al.(2021)Mehrabi, Morstatter, Saxena, Lerman, and Galstyan]{mehrabi2021survey}
Ninareh Mehrabi, Fred Morstatter, Nripsuta Saxena, Kristina Lerman, and Aram Galstyan.
\newblock A survey on bias and fairness in machine learning.
\newblock \emph{ACM computing surveys (CSUR)}, 54\penalty0 (6):\penalty0 1--35, 2021.

\bibitem[Nam et~al.(2020)Nam, Cha, Ahn, Lee, and Shin]{nam2020learning}
Junhyun Nam, Hyuntak Cha, Sungsoo Ahn, Jaeho Lee, and Jinwoo Shin.
\newblock Learning from failure: De-biasing classifier from biased classifier.
\newblock \emph{Advances in Neural Information Processing Systems}, 33:\penalty0 20673--20684, 2020.

\bibitem[Park et~al.(2023)Park, Lee, and Ye]{park2023efficient}
Geon~Yeong Park, Sang~Wan Lee, and Jong~Chul Ye.
\newblock Efficient debiasing with contrastive weight pruning, 2023.
\newblock URL \url{https://openreview.net/forum?id=0DIkhwclYX3}.

\bibitem[Qraitem et~al.(2023)Qraitem, Saenko, and Plummer]{qraitem2023bias}
Maan Qraitem, Kate Saenko, and Bryan~A Plummer.
\newblock Bias mimicking: A simple sampling approach for bias mitigation.
\newblock In \emph{Proceedings of the IEEE/CVF Conference on Computer Vision and Pattern Recognition}, pages 20311--20320, 2023.

\bibitem[Sagawa et~al.(2019)Sagawa, Koh, Hashimoto, and Liang]{sagawa2019distributionally}
Shiori Sagawa, PangWei Koh, Tatsunori~B Hashimoto, and Percy Liang.
\newblock Distributionally robust neural networks for group shifts: On the importance of regularization for worst-case generalization.
\newblock \emph{arXiv preprint arXiv:1911.08731}, 2019.

\bibitem[Selvaraju et~al.(2017)Selvaraju, Cogswell, Das, Vedantam, Parikh, and Batra]{selvaraju2017grad}
Ramprasaath~R Selvaraju, Michael Cogswell, Abhishek Das, Ramakrishna Vedantam, Devi Parikh, and Dhruv Batra.
\newblock Grad-cam: Visual explanations from deep networks via gradient-based localization.
\newblock In \emph{Proceedings of the IEEE international conference on computer vision}, pages 618--626, 2017.

\bibitem[Tartaglione et~al.(2021)Tartaglione, Barbano, and Grangetto]{tartaglione2021end}
Enzo Tartaglione, Carlo~Alberto Barbano, and Marco Grangetto.
\newblock End: Entangling and disentangling deep representations for bias correction.
\newblock In \emph{Proceedings of the IEEE/CVF conference on computer vision and pattern recognition}, pages 13508--13517, 2021.

\bibitem[Teney et~al.(2020)Teney, Abbasnejad, and van~den Hengel]{teney2020unshuffling}
Damien Teney, Ehsan Abbasnejad, and Anton van~den Hengel.
\newblock Unshufﬂing data for improved generalization.
\newblock \emph{arXiv preprint arXiv:2002.11894}, 2020.

\bibitem[Torralba and Efros(2011)]{torra}
Antonio Torralba and Alexei~A. Efros.
\newblock Unbiased look at dataset bias.
\newblock In \emph{CVPR 2011}, pages 1521--1528, 2011.
\newblock \doi{10.1109/CVPR.2011.5995347}.

\bibitem[Vandenhirtz et~al.(2023)Vandenhirtz, Manduchi, Marcinkevi{\v{c}}s, and Vogt]{vandenhirtz2023signal}
Moritz Vandenhirtz, Laura Manduchi, Ri{\v{c}}ards Marcinkevi{\v{c}}s, and Julia~E Vogt.
\newblock Signal is harder to learn than bias: Debiasing with focal loss.
\newblock \emph{arXiv preprint arXiv:2305.19671}, 2023.

\bibitem[Wang et~al.(2019)Wang, He, Lipton, and Xing]{wang2019learning}
Haohan Wang, Zexue He, Zachary~C Lipton, and Eric~P Xing.
\newblock Learning robust representations by projecting superficial statistics out.
\newblock \emph{arXiv preprint arXiv:1903.06256}, 2019.

\bibitem[Wu et~al.(2023)Wu, Yuksekgonul, Zhang, and Zou]{wu2023discover}
Shirley Wu, Mert Yuksekgonul, Linjun Zhang, and James Zou.
\newblock Discover and cure: Concept-aware mitigation of spurious correlation.
\newblock In \emph{ICML}, 2023.

\bibitem[Zhang and Sabuncu(2018)]{zhang2018generalized}
Zhilu Zhang and Mert Sabuncu.
\newblock Generalized cross entropy loss for training deep neural networks with noisy labels.
\newblock \emph{Advances in neural information processing systems}, 31, 2018.

\end{thebibliography}
\end{document}

\maketitle
\begin{abstract}
In this supplementary material, we provide additional results on the Biased Action Recognition (BAR) dataset and more details about the other datasets used. We include results on the effect of the hyperparameter $\lambda_c$ on model performance. Additionally, we examine the performance of our model on bias-aligned and bias-conflicting samples separately. We explore the impact of layer selection on model performance and characterize the effects of applying similarity versus dissimilarity. Furthermore, we provide Grad-CAM visualizations comparing our model to the vanilla model. Finally, we include a detailed algorithm of our model architecture.
   
\end{abstract}

\section{Dataset Description}
We assess our debiasing algorithm on four standard datasets. (1) The \textbf{Colored MNIST} dataset is the modified version of the MNIST dataset~\cite{lecun2010mnist} where each digit is deliberately perturbed by a specific color to form a spurious correlation between bias attributes and target labels. Instead of creating our own version of the dataset, we use the Colored MNIST dataset as proposed in \cite{nam2020learning}  to ensure a fair and direct comparison. This is because the dataset's construction significantly varies due to the changing coloring protocol.  (2) The \textbf{Corrupted CIFAR-$10^1$}~\cite{hendrycks2019benchmarking} dataset introduces bias attributes through artificially generated distortions, such as fog, adjustments in brightness, or variations in saturation, which are synthetically applied to each class. (3) The \textbf{BFFHQ dataset}~\cite{kim2021biaswap} is a gender-biased derivative of the widely-used Flickr-Faces-HQ (FFHQ) dataset, where age serves as the target label and gender is a correlated bias. (4) The \textbf{Dogs $\&$ Cats dataset}~\cite{kim2019learning} is an (animal, color) dataset, where the former and latter represent target and bias attributes, respectively. Each dataset is predominantly composed of biased samples (up to $99.5\%$), with a minimal presence of bias-conflicting samples. The distribution of bias samples ranges from $95\%$ to $99.5\%$. 

The dimensions of the images differ across the datasets: Colored MNIST has dimensions of 28×28, Corrupted-CIFAR-$10^1$ has dimensions of 32x32, and the remaining datasets have dimensions of 224×224. All image data is normalized across each channel using a mean of (0.4914, 0.4822, 0.4465) and a standard deviation of (0.2023, 0.1994, 0.2010). While Colored MNIST does not undergo additional data augmentation, all other datasets benefit from random crop and horizontal flip transformations
\vspace{1em}

\noindent \textbf{Evaluation Protocol:} We follow the standard protocol for evaluation as used in the other SOTA methods, where the train and validation sets are similar while the test set is inverted. For example, in training (99\% BA dataset and 1\% BC) while in testing (99\%BC and 1\% BA)
\noindent Our idea is based on the notion of learning non-spurious and inherent features of objects by imposing both dissimilarity constraints in the final layers and similarity in the initial layers. However, for a comprehensive ablation study, we will also include results in below table obtained using only similarity constraints.

\vspace{-2mm}\subsection{BFFHQ} The BFFHQ dataset \cite{kim2021biaswap} is a gender-biased derivative of the widely-used Flickr-Faces-HQ (FFHQ) dataset, with age as the target label and gender as a correlated bias. It predominantly comprises the 'young' category (individuals aged 10 to 29) which is strongly associated with the 'female' gender, while the 'old' category (individuals aged 40 to 59) is linked with the 'male' gender.

\vspace{-2mm}\subsection{Dogs \& Cats}
The Dogs \& Cats dataset, initially presented by \cite{kim2019learning} and subsequently restructured by \cite{Lee2022} for bias studies from the original training set, features images categorized into "bright dogs and dark cats" versus "dark dogs and bright cats". Within this dataset, the proportion of bias-aligned to bias-conflicting samples is determined as 8037 to 80, which corresponds to a 1\% ratio, and 8037 to 452, equating to a 5\% ratio. According to \cite{ritter2017cognitive}, neural networks show a predilection for learning shapes over colors, provided the dataset is sufficiently robust. In this context, the dataset stands out from the Colored MNIST dataset in that the shapes are more pronounced, while the color of the animals is de-emphasized.

\vspace{-3mm}\subsection{Colored MNIST} A modified version of MNIST dataset \cite{lecun2010mnist} where we pertubate them systematically with colours that act as bias attributes. Each of the ten digits is intentionally correlated with a particular color (e.g., red for digit 0). To materialize this bias, color is strategically injected into the foreground of each image. Our experimentation with the Colored MNIST dataset, following the methodology established in  \cite{Lee2022}, involves a systematic consideration of different ratios of bias-conflicting samples. For each ratio of interest, the dataset is divided into two subsets: bias-aligned samples and bias-conflicting samples. The distribution of these subsets varies depending on the specific ratio being examined. Here are the exact numbers for each ratio:
\begin{itemize}
\itemsep=-0.5em
    \item 0.5\% ratio: 54751 bias-aligned samples, 249 bias-conflicting samples
    \item 1\% ratio: 54509 bias-aligned samples, 491 bias-conflicting samples
    \item 2\% ratio: 54014 bias-aligned samples, 986 bias-conflicting samples
    \item 5\% ratio: 52551 bias-aligned samples, 2449 bias-conflicting samples.
\end{itemize}
This deliberate configuration allows us to comprehensively explore and analyze the impact of bias and its conflicts on the performance of machine learning models in the context of Colored MNIST.

\vspace{-3mm}\subsection{Corrupted CIFAR-$10^1$} The second dataset is the perturbed version of original \cite{krizhevsky2009learning}. We adopted the version used by \cite{hendrycks2019benchmarking}, where the data is perturbed with artificially generated distortions, such as fog, adjustments in brightness, or variations in saturation, have been synthetically introduced to each class. These carefully crafted synthetic biases aim to mimic real-world scenarios closely.

\vspace{-3mm}
\subsection{BAR}
The Biased Action Recognition (BAR) dataset comprises six action classes, each biased toward distinct environmental contexts. These classes were chosen by examining the imSitu dataset(a dataset supporting situation recognition), which provides action and place labels for still images from Google Image Search. The selected action classes share commonplace characteristics while having distinct place attributes, resulting in six typical action-place pairs: (Climbing, Rock Wall), (Diving, Underwater), (Fishing, Water Surface), (Racing, A Paved Track), (Throwing, Playing Field), and (Vaulting, Sky).

\section{ Results on BAR dataset}
In our experiments with the BAR dataset presented in Table ~\ref{table:bar_results}, the anticipated performance levels were not achieved. One plausible explanation for this outcome can be traced back to the limited size of the dataset. Specifically, under the 99\% bias ratio setting for class "0", there are only two images that correspond to bias-conflicting samples, in stark contrast to the 296 bias-aligned images. The constraints imposed by the diminutive dataset size necessitated the use of pre-trained models. Consequently, our model faces significant limitations in its learning capacity under such conditions, where bias-conflicting samples are extremely scarce. Given all these limitations, our model still gives better performance than baseline \cite{nam2020learning} as well as recent studies carried out by \cite{amplibias,Lee2022, an20222}  on 99\% bias ratio. However, when evaluated on a 95\% bias ratio, we outperform the baseline, although our model's accuracy falls short of the margin in accommodating new developments.

\begin{table}[h]
\centering
% \small
 % \footnotesize
\begin{tabular}{|c|c|c|}
\hline
& \multicolumn{2}{c|}{BAR Dataset \cite{nam2020learning}} \\
\hline  \hline
\textbf{Bias Ratio(\%)} & \makebox[1.5cm]{99.00} & \makebox[1.5cm]{95.00} \\ \hline
\hline
Vanilla & 70.55 & 82.53 \\
\hline
DisEnt \cite{lee2021learning} & 70.33 & 83.13 \\
\hline
LFF \cite{nam2020learning} & 70.16 & 82.95 \\
\hline
A$^2$ \cite{an20222} & 71.15 & 83.07 \\  
\hline
ReBias \cite{bahng2020learning} & 73.01 & 83.51  \\
\hline
Revisiting LFF \cite{Lee2022}& 73.36 & \textbf{84.96} \\
\hline
AmpliBias \cite{amplibias} & 73.30  & 84.67 \\
\hline
Ours & \textbf{74.52} & 83.48 \\
\hline
% Ours with Dissim & 70.12 & 83.83 \\
% \hline 
\end{tabular}
\caption{Comparative accuracy performance (in \%) of of various
debiasing algorithms on the BAR Dataset.}
\label{table:bar_results}
% \vspace{-2em}
\end{table}

\section{Effect of hyperparameter $\lambda_c$}

The hyper-parameters can have a profound impact on model performance and need to be adjusted accordingly to get the intended performance. In our method, we adjust the strength of similarity or dissimilarity, with respect to where it is applied to the $F_d$ model using $\lambda_c$. This adjustment is necessary because of some intrinsic factors. Since we are making the layers of bias $F_d$ and debias $F_d$ model dissimilar, it may not always be the case that they need to be completely similar or completely dissimilar. The bias model can have useful information that should be retained by the $F_d$ model to perform well. In fact, to make enhanced predictions on both bias-aligned images, the debias model should also have qualities of the bias model as the debias model is already best in giving the highest performance on bias-aligned images.

For the Coloured MNIST dataset, we see that the highest performance is achieved when $\lambda_c$ is equal to $0.1$. In the case of coloured MNIST, the bias is easy and is learned at early layers; since it is easy it also impacts the $F_d$ model easily. Hence, to debias we need to have a strong $\lambda_c$ to control the bias and make it dissimilar from $F_b$. On the other hand, in BFFHQ Dataset, the dissimilarity strength is very low (between $10^{-5} \text{ to } 10^{-8}$). The reason can be attributed to the presence of similar high-level features in both $F_d$ and $F_b$. Since, the high-level abstractions are very similar for both the models, making them very dissimilar is bound to distort the performance of the model. The same pattern can be seen in Cats \& Dogs Dataset where dissimilarity strength is $10^{-8}$ for a very severe 99\% bias aligned dataset and $10^{-3}$ for 95\% biased dataset. 
% In addition to these, for Corrupted CIFAR-$10^1$ dataset, the value of $\lambda_c$ is low to moderate for different bias conflicting ratios. 
\begin{table}[h]
\centering
% \small
\begin{tabular}{|c|c|c|c|c|}
\hline
& \multicolumn{4}{c|}{CMNIST Dataset} \\ \hline \hline
$\lambda_c$ & 99.50 & 99.00 & 98.00 & 95.00\\ \hline
$10^{-1}$ & \textbf{66.98} & \textbf{78.03} & \textbf{84.22} & \textbf{88.64} \\ \hline
$10^{-2}$ & 60.22 & 73.41 & 81.12 & 86.37 \\ \hline
$10^{-3}$ & 62.41 & 74.14 & 78.36 & 84.84 \\ \hline
$10^{-4}$ & 60.81 & 74.57 & 79.43 & 85.81 \\ \hline
$10^{-5}$ & 64.36 & 75.18 & 80.02 & 84.48 \\ \hline
$10^{-7}$ & 63.83 & 76.29 & 79.51 & 84.99 \\ \hline
$10^{-8}$ & 64.57 & 77.10 & 80.04 & 84.39 \\ \hline
\end{tabular}
\caption{Performance Impact of Varying Dissimilarity Strength $\lambda_c$ on the Third Layer of a 3-Layer MLP Model on the CMNIST Dataset. The table presents accuracy (in \%) across different bias ratios in the dataset, illustrating the effect of $\lambda_c$ on robustness against bias.}
\label{Table:lambda_c_CMNIST}
\end{table}

\begin{table}[h]
\centering
    % \small
    \begin{tabular}{|c|cccc|}
        \hline
        & \multicolumn{4}{c|}{BFFHQ Dataset} \\ \hline \hline
        $\lambda_c$ & \multicolumn{1}{c|}{99.50} & \multicolumn{1}{c|}{99.00} & \multicolumn{1}{c|}{98.00} & 95.00 \\ \hline
        $10^{-1}$ & \multicolumn{1}{c|}{82.20} & \multicolumn{1}{c|}{78.80} & \multicolumn{1}{c|}{88.40} & 86.40 \\ \hline
        $10^{-2} $& \multicolumn{1}{c|}{81.60} & \multicolumn{1}{c|}{80.00} & \multicolumn{1}{c|}{86.40} & 89.20 \\ \hline
        $10^{-3}$ & \multicolumn{1}{c|}{82.20} & \multicolumn{1}{c|}{78.40} & \multicolumn{1}{c|}{87.00} & 88.60 \\ \hline
        $10^{-4}$ & \multicolumn{1}{c|}{79.80} & \multicolumn{1}{c|}{80.00} & \multicolumn{1}{c|}{88.40} & 88.80 \\ \hline
        $10^{-5} $& \multicolumn{1}{c|}{81.40} & \multicolumn{1}{c|}{79.50} & \multicolumn{1}{c|}{88.00}  & \textbf{90.20} \\ \hline
        $10^{-6}$ & \multicolumn{1}{c|}{\textbf{83.20}} & \multicolumn{1}{c|}{\textbf{82.20}} & \multicolumn{1}{c|}{86.80} & 88.60 \\ \hline
        $10^{-8} $& \multicolumn{1}{c|}{81.40} & \multicolumn{1}{c|}{78.70} & \multicolumn{1}{c|}{\textbf{88.40}} & 87.00 \\ \hline
    \end{tabular}
    \caption{ Performance Impact of Varying Dissimilarity Strength $\lambda_c$ on the second MLP layer of pre-trained ResNet18 with three MLP layers on the BFFHQ dataset. The last MLP layer corresponds to the classification layer. The table presents accuracy (in \%) across different bias ratios in the dataset, illustrating the effect of $\lambda_c$ on model robustness against bias.}
    \label{Table:lambda_c_BFFHQ}
\end{table}

\begin{table}[h]
\centering
% \small
    \begin{tabular}{|c|c|c|}
        \hline
        & \multicolumn{2}{c|}{Cats \& Dogs Dataset} \\ \hline \hline
        $\lambda_c$ & \makebox[1.5cm]{99.00} & \makebox[1.5cm]{95.00} \\ \hline
        $10^{-1}$ & 90.40 & 95.00 \\ \hline
        $10^{-2}$ & 89.90 & 95.60 \\ \hline
        $10^{-3}$ & 89.80 & \textbf{96.50} \\ \hline
        $10^{-4}$ & 89.20 & 94.10 \\ \hline
        $10^{-5}$ & 90.70 & 94.10 \\ \hline
        $10^{-6}$ & 92.20 & 94.80 \\ \hline
        $10^{-8}$ & \textbf{93.00} & 95.00 \\ \hline
    \end{tabular}
    \caption{ Performance with varying  $\lambda_c$ on the second MLP layer of pre-trained ResNet18 with three MLP layers on the Cats \& Dogs dataset. The table presents accuracy (in \%) across different bias ratios in the dataset.}
    \label{Table:lambda_c_CATSDOGS}
\end{table}
% \end{wraptable}

% \section{Why parameters? }
% We agree that mitigating bias in either the feature or sample space is a commendable approach. However, our method offers a potential direction for debiasing by utilising the parameter space. By combining both feature space and parameter space, it is possible to achieve improved accuracy in debiasing biased datasets.

\section{Effect on bias aligned and bias conflicting Samples}
\begin{figure}[h]
    \centering
    \includegraphics[scale=0.4]{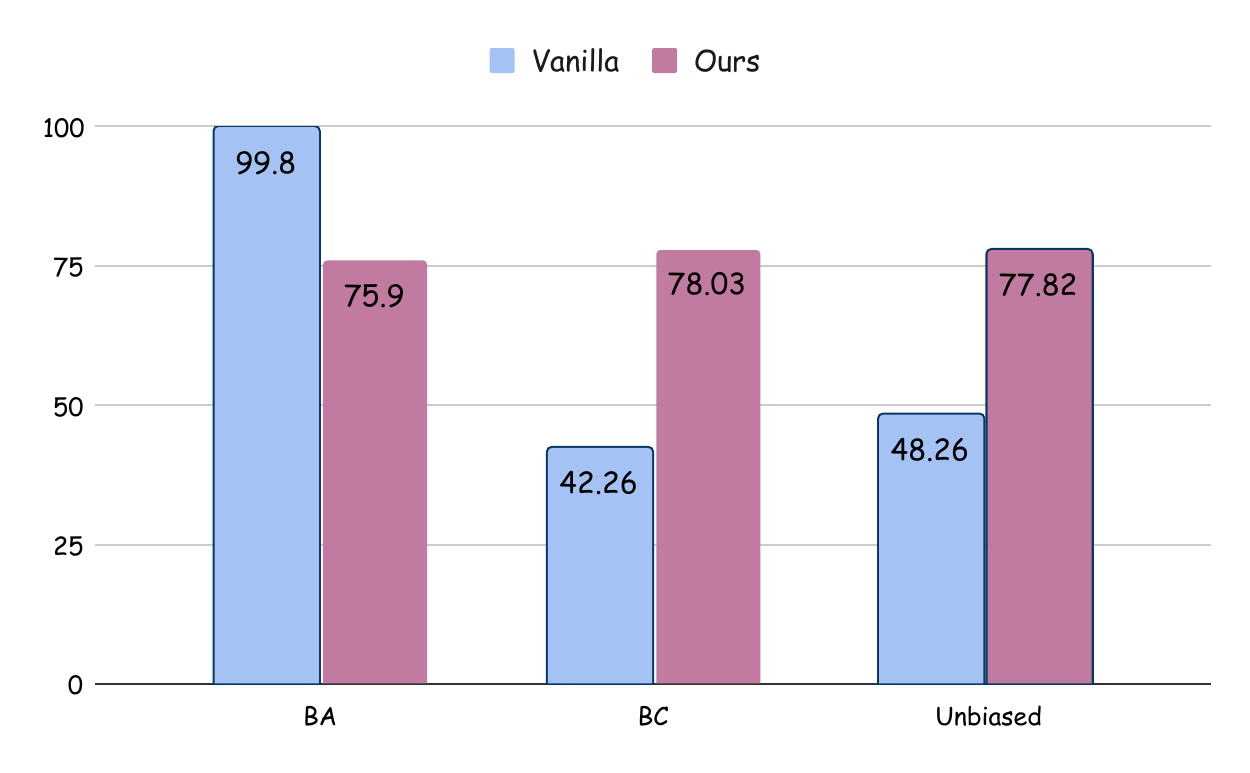}
    \caption{Comparative Analysis of Model Performance on Bias-Aligned, Unbiased, and Bias-Conflicting Samples. In this figure, 'BA' denotes Bias-Aligned, and 'BC' denotes Bias-Conflicting samples. The unbiased dataset has images from both the BA and BC sets.}
    \label{fig:BA_BC_SUPP}
\end{figure}

Our model -CosFairNet demonstrates superior performance on both unbiased and bias-conflicting samples in comparison to the vanilla model. The latter tends to acquire biases, which are simpler to assimilate than the true underlying signals, and thus, it predominantly relies on these biases for classification. As depicted in Figure \ref{fig:BA_BC_SUPP}, the vanilla model exhibits near-perfect accuracy in recognizing bias-aligned samples. However, it significantly underperforms in accurately classifying bias-conflicting and overall unbiased samples, highlighting the effectiveness of our approach in mitigating bias. Furthermore, it is important to note that classifying bias-aligned samples is not as critical as identifying bias-conflicting samples. Therefore, despite the vanilla model performing exceptionally well on bias-aligned samples, its practical value is limited.

\section{Effect of layer selection}

Biases come in different magnitudes, with some being easy to address and others more challenging. Depending on the ease of bias within the dataset, they are learned at either the early layers, mid-layers, or the final layers of the deep learning model. In Table ~\ref{Table:layer_selection}, we present an ablation study of applying similarity or dissimilarity at different layers of the model. We observe a few things - first, applying layer similarity at just one layer works very well and yields optimal results, whereas applying it to multiple layers leads to a drastic degradation in performance. For instance, applying similarity or dissimilarity to the second layer of a three-layer dense network results in very high accuracy, surpassing the present state-of-the-art. However, when applied to both the first and second layers, the performance drops to less than half of the intended accuracy. Additionally, controlling bias at earlier layers seems to be generally more effective compared to controlling bias at later layers of the model.

\begin{table}[h]
\centering
% \scriptsize
\begin{tabular}{|c|ccccccc|}
\hline
 &
  \multicolumn{7}{c|}{CMNIST} \\ \hline \hline
 &
  \multicolumn{1}{c|}{\textbf{1}} &
  \multicolumn{1}{c|}{\textbf{2}} &
  \multicolumn{1}{c|}{\textbf{3}} &
  \multicolumn{1}{c|}{\textbf{12}} &
  \multicolumn{1}{c|}{\textbf{13}} &
  \multicolumn{1}{c|}{\textbf{23}} &
  \textbf{123} \\ \hline
Similarity &
  \multicolumn{1}{c|}{74.3} &
  \multicolumn{1}{c|}{75.13} &
  \multicolumn{1}{c|}{73.83} &
  \multicolumn{1}{c|}{30.05} &
  \multicolumn{1}{c|}{41.24} &
  \multicolumn{1}{c|}{69.85} & 26.76
   \\  \hline
Dissimilarity &
  \multicolumn{1}{c|}{65.11} &
  \multicolumn{1}{c|}{\textbf{78.03}} &
  \multicolumn{1}{c|}{76.31} &
  \multicolumn{1}{c|}{28.45} &
  \multicolumn{1}{c|}{30.57} &
  \multicolumn{1}{c|}{72.54} & 27.94
   \\ \hline
   % \vspace{-0.4em}
\end{tabular}

\caption{Performance Comparison of applying Similarity and Dissimilarity on Different Layers of a 3-Layer MLP model on the CMNIST Dataset with 99\% Bias Ratio. This table represents the impact of applying similarity and dissimilarity metrics to the first (1), second (2), and third (3) layers, as well as their combinations (12, 13, 23, 123), elucidating how interventions in specific layers affect the overall model performance.}
\label{Table:layer_selection}
\end{table}

\vspace{1em}
\par
\begin{figure}[h]
\centering
    \includegraphics[scale=0.6]{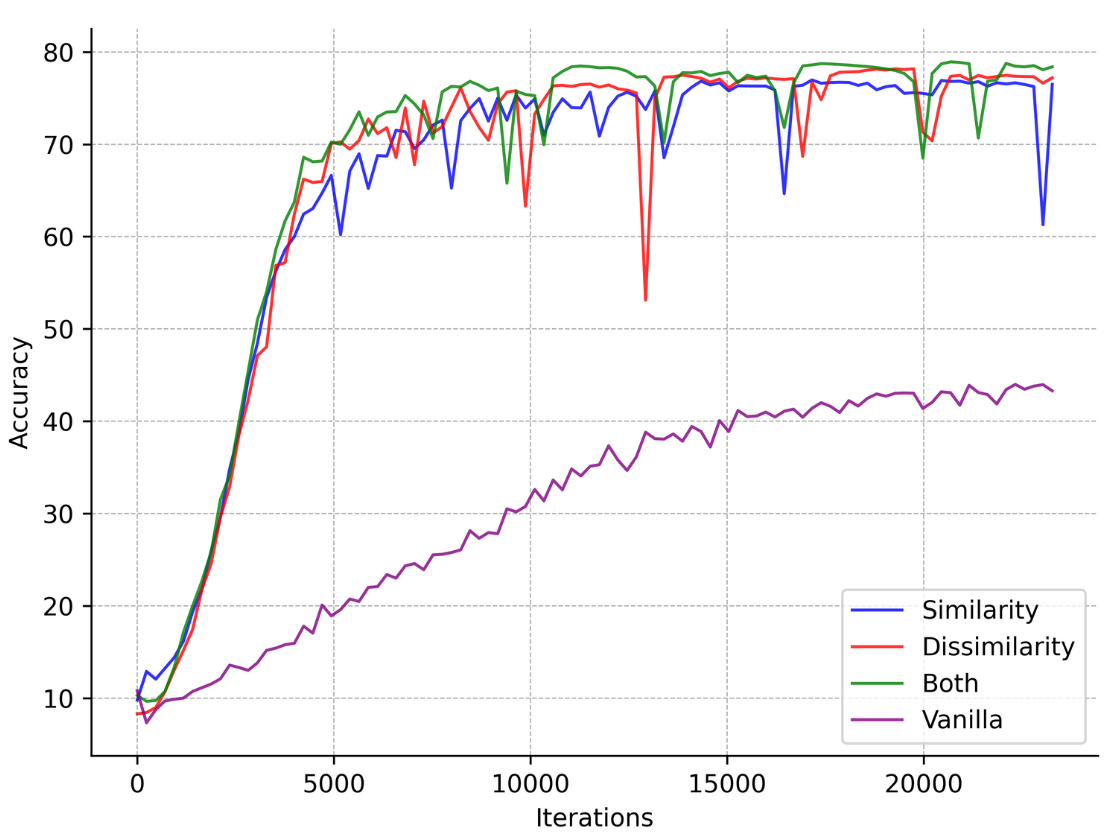}
    \caption{ Performance comparison of our method against baselines with increasing training iteration. The results are shown  on the CMNIST dataset that has a 99\% bias ratio.``Both" refers to our final model after the combined application of similarity and dissimilarity constraints. (Best view in colour).}
    \label{fig:acc_vs_it}
\end{figure}

 Figure~\ref{fig:acc_vs_it} demonstrates our model's performance improvement over time, especially when employing both similarity and dissimilarity constraints, surpassing other models in effectiveness and convergence. This highlights the importance of layer-specific orthogonality in strengthening debiasing efforts. 

% \section{Statistical Test}

% \begin{figure}[H]
%     \centering
%     \includegraphics[width=0.75\linewidth]{BFFHQ.png}
%     \caption{Critical Difference (CD) Diagram for Model Comparison on the BFFHQ Dataset (99\% bias ratio).}
%     \label{fig:enter-label}
% \end{figure}

% \begin{figure}[H]
%     \centering
%     \includegraphics[width=0.75\linewidth]{cmnist_stat.png}
%     \caption{Critical Difference (CD) Diagram for Model Comparison on the Coloured MNIST Dataset (99\% bias ratio)}
%     \label{fig:enter-label}
% \end{figure}

\section{Grad-CAM Visualization}
\begin{figure}[h]
    \centering
    \includegraphics[scale=0.65]{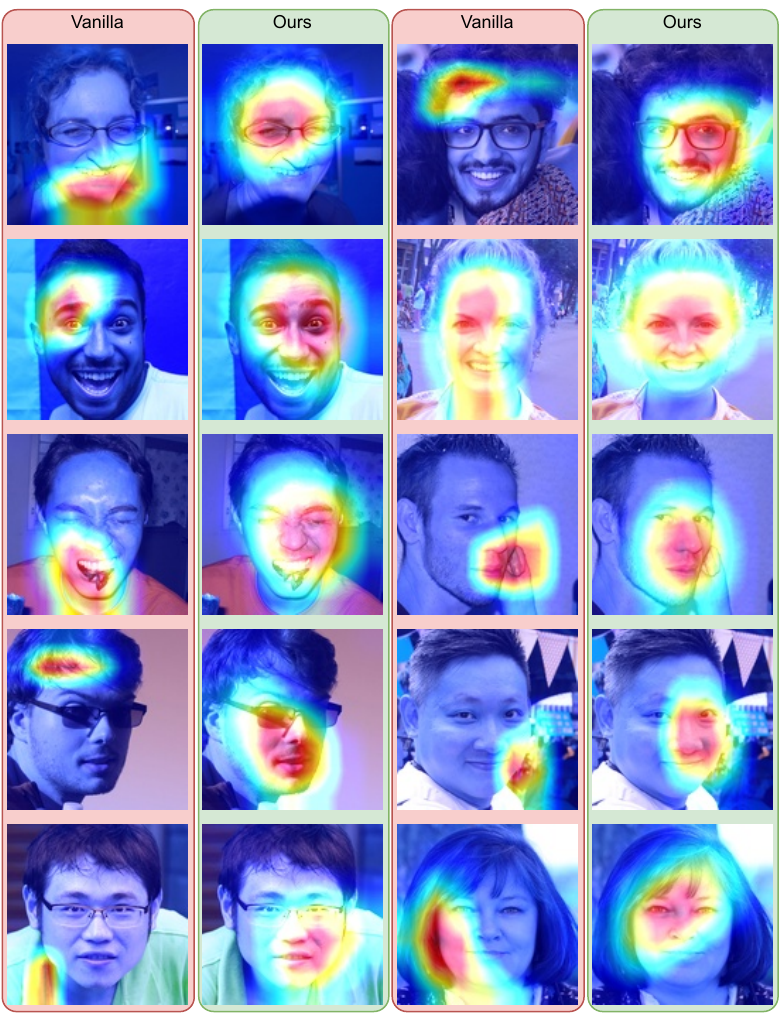}
    \caption{Grad-CAM Visualization Comparing Vanilla and Our Model's Attention on the BFFHQ Dataset.  The left column, 'Vanilla', shows a concentration on gendered features while 'Ours' on the right demonstrates a more equitable distribution of attention, emphasizing our model's resistance to gender bias.}
    \label{fig:grad_cam_bbfhq_supp}
\end{figure}

In Figure~\ref{fig:grad_cam_bbfhq_supp}, using Grad-CAM visualizations, we present a comparative analysis between our proposed model- CosFairNet and the vanilla model in the context of the BFFHQ dataset \cite{kim2021biaswap}. The GradCam visualizations elucidate the regions within the images that the models focus on to infer age, which serves as the target label, amidst the gender-biased data. It is evident from the 'Vanilla' model columns that the attention is unevenly distributed across gender lines. On the other hand, our model demonstrates a more uniform attention distribution, indicating a generalized approach towards learning gender when determining age. This is particularly noteworthy in images where the 'old' category overlaps with 'female' characteristics and the 'young' with 'male', showcasing our model's improved capability to generalize beyond the biased associations present in the dataset.

\section{Algorithm}
The proposed algorithm \ref{algo}, CosFairNet, aims to train a debiased model by leveraging cosine similarity to realign model parameters. The training process involves a biased model \(F_b(x; \theta_b)\) and a debiasing model \(F_d(x; \theta_d)\), both initialized with their respective parameters. For each mini-batch of labeled samples, predictions are generated from both models, followed by the computation of cross-entropy loss for the debiasing model and generalized cross-entropy loss for the biased model. The biased model parameters are updated using gradient descent. The difficulty score is then obtained to weight the loss for updating the debiasing model. Additionally, a dissimilarity loss is introduced, which uses cosine similarity to ensure the initial layers of both models are aligned while the later layers diverge. This alignment aims to reduce bias by penalizing similarity in later layers, thereby promoting diversity in learned features. The algorithm iteratively updates the debiased model parameters, ultimately yielding a model with minimized bias and enhanced generalization.

\begin{algorithm*}[!]
\SetAlgoLined
\KwIn{
    Training data $\mathcal{D}=\left\{(x_{1}, 
y_1),...,(x_N,y_N)\right\}$ 
\\ Parameters $\theta_b$ and $\theta_d$,  learning rate $\eta$,  cosine strength $\lambda_c$, Initialize biased $F_b (x; \theta_b)$ and debiasing $F_d (x; \theta_d)$ model\\
}
\KwOut{Trained debiased model $F(;\hat{\hat{\theta}}_d)$}
\For{mini batch of labeled samples $(x_i, y_i)\in \mathcal{D}$}{
Prediction from bias model: 
$\hat{y}^b_i \leftarrow$ $F_b(x_i,\theta_b)$\; 
Prediction from debias model:  
$\hat{y}^d_i \leftarrow$ $F_d(x_i,\theta_d)$\;

Losses: 
$ \mathcal{L}_{d} $ $\leftarrow$   $\operatorname{CE} (\hat{y}^d_i;y_i) $ \\
\hspace{33pt}$ \mathcal{L}_{b} $ $\leftarrow$   $\operatorname{GCE} (\hat{y}^b_i;y_i) $  \\
\textbf{Update the biased model: }\\
 $\hat{\theta_b}$ $\leftarrow$ $\theta_b$ -$\eta\frac{\partial\mathcal{L}_{b}}{\partial \theta_b}$ \\
 Obtain  $ \mathcal{W}(x_i)$ from Eq.~\ref{eq:difficulty_score} \\
 \textbf{Update the debiased model: }\\
  $\hat{\theta_d}$ $\leftarrow$ $\theta_d$ -$\eta\frac{\partial \mathcal{W}(x_i) \cdot \mathcal{L}_{d}}{\partial \theta_d}$ \\
\textbf{Calculate the dis-similarity loss: } \\
\For{$k^{th}$ layer parameters $\hat{\theta}_{b(k)}\in \hat{\theta}_b$ and  $\hat{\theta}_{d(k)}\in \hat{\theta}_d$}{

 \uIf{$k$ is initial layer}{
      $  \lambda_c\mathcal{L}_{cosSim}(\hat{\theta}_{d(k)},\hat{\theta}_{b(k)})$ from Eq.~\ref{eq:cosine_loss} \\
  $\hat{\hat{\theta}}_d$ $\leftarrow$ $\hat{\theta_d}$ -$\eta\frac{\partial \mathcal{L}_{cosSim}}{\partial \hat{\theta}_d}$  \;
  }
   \Else{
   $ \mathcal{L}_{dis} $ $\leftarrow$    $ \lambda_c \left ( 1 - \mathcal{L}_{cosSim}(\hat{\theta}_{d(k)},\hat{\theta}_{b(k)}) \right ) $  \\
  $\hat{\hat{\theta}}_d$ $\leftarrow$ $\hat{\theta_d}$ -$\eta\frac{\partial \mathcal{L}_{dis}}{\partial \hat{\theta}_d}$  \;
  }
 
}
}
% Output debiased models with parameters $\hat{\hat{\theta}}_d$
\caption{CosFairNet: Parameters Realignment using Cosines}
\label{algo}
% \vspace{-0.5em}
\end{algorithm*}

\vspace{1em}

\noindent The code for the proposed model \enquote{CosFairNet} and additional details can be found on the project page: \href{https://visdomlab.github.io/CosFairNet/}{https://visdomlab.github.io/CosFairNet/}

\newpage

\bibliography{egbib}